# Aggregating Learned Probabilistic Beliefs

**Pedrito Maynard-Reid II**
Computer Science Department
Stanford University
Stanford, CA 94305-9010
pedmayn@cs.stanford.edu

**Urszula Chajewska**
Computer Science Department
Stanford University
Stanford, CA 94305-9010
urszula@cs.stanford.edu

## Abstract

We consider the task of aggregating beliefs of several experts. We assume that these beliefs are represented as probability distributions. We argue that the evaluation of any aggregation technique depends on the semantic context of this task. We propose a framework, in which we assume that nature generates samples from a 'true' distribution and different experts form their beliefs based on the subsets of the data they have a chance to observe. Naturally, the optimal aggregate distribution would be the one learned from the combined sample sets. Such a formulation leads to a natural way to measure the accuracy of the aggregation mechanism.

We show that the well-known aggregation operator LinOP is ideally suited for that task. We propose a LinOP-based learning algorithm, inspired by the techniques developed for Bayesian learning, which aggregates the experts' distributions represented as Bayesian networks. We show experimentally that this algorithm performs well in practice.

## 1 Introduction

Belief aggregation of subjective probability distributions has been a subject of great interest in statistics (see [GZ86, CW99]) and, more recently, artificial intelligence (e.g., [PW99]) and machine learning (ensemble learning in particular [PMGH00]), especially since probabilistic distributions are increasingly being used in medicine and other fields to encode knowledge of experts. Unfortunately, many of the aggregation proposals have lacked sufficient semantical underpinnings, typically evaluating a mechanism by how well it satisfies properties justified by little more than intuition. However, as has been noted in other fields such as belief revision (cf. [FH96]), the appropriateness of properties depends on the particular context.

We take a more semantic approach to aggregation: we first describe the realistic framework in which the experts or *sources* learn their probability distributions from data using standard probabilistic learning techniques. We assume a Decision Maker (DM) — the traditional name for the aggregator — wants to aggregate a set of these learned distributions. This framework suggests a natural optimal aggregation mechanism: construct the distribution that would be learned had all the sources' data sets been available to the DM. Since the original data sets are generally not available, the aggregation mechanism should come as close as possible to reconstructing the data sets and learning from the combined set.

For intuition, consider the the task of creating an expert system for some specialized medical field. We would like to take advantage of the expertise of several doctors working in this field. Each of these doctors sharpened his knowledge by following many patients. The doctors can no longer recall the specifics of each case, but they have formed over the years fairly accurate models of the domain that can be represented as sets of conditional probabilities. (In fact, many expert systems have been created over the years by eliciting such conditional probabilities from experts [HHN92].) Of course, if there was a doctor who had seen all of the patients the others doctors saw, the ideal expert system would result from eliciting her model. However, there isn't one such expert. Therefore, our system would benefit from incorporating the knowledge of as many experts as we can find. The system would also account for the differing levels of experience of different doctors -- some of them may have practiced for much longer than others.

One of the best-known aggregation operators is the Linear Opinion Pool (LinOP) which aggregates a set of distributions by taking their weighted sum. It has been shown in the statistics community that, under some intuitive assumptions, learning the joint distribution from the combined data set is equivalent to using LinOP over the individual joint distributions learned from the individual data sets. However, whereas the weights in typical uses of LinOP are often criticized for being ad-hoc, our framework prescribes semantically-justified weights: the estimated percentages of the data each source saw. Intuitively, a high weight means we believe a source has seen a relatively large amount of data and is, hence, likely to be reliable.

However, joint distributions are hardly the preferred representation for probabilistic beliefs in real-world domains. BNs (aka belief networks, etc.) [Pea88] have gained much popularity as structured representations of probability distributions. They allow such distributions to be represented much more compactly, therefore often avoiding exponential blowup in both memory size and inference complexity.

Thus, we assume the sources beliefs are BNs learned from data. According to our semantics, the aggregate BN should be one the DM would learn from the combined sets of data. We describe a LinOP-based BN aggregation algorithm, inspired by the algorithm designed to learn BNs from data. The algorithm uses sources' distributions instead of samples to search over possible BN structures and parameter settings. It takes advantage of the marginalization property of LinOP to make computation more efficient. We explore the algorithm's behavior by running experiments on the well-known, real-life Alarm network [BSCC89] and on the smaller artificial Asia network [LS88].

## 2 Formal Preliminaries

We restrict our attention to domains with discrete variables. We consider how to compute the aggregate distribution, and how the accuracy of our computation depends on how much we know about the sources.

Formally, we consider the following setting: There are $L$ sources and $N$ discrete random variables, where each variable $X$ has domain dom(X). We follow the convention of using capital letters to denote variables and lowercase letters to denote their values. Symbols in bold denote sets. $\mathcal{W}$ is the set of possible worlds defined by value assignments to variables. The true distribution or model of the world is $\pi$. Each source $i$ has a data set $\mathbf{D}_i$ sampled from (unknown to us) $\pi$. We will assume that each $\mathbf{D}_i$ is finite of size $M_i$. The corresponding empirical (i.e., frequency) distribution is $\hat{p}_i$. Each source $i$ learns a distribution $p_i$ over $\mathcal{W}$. This is $i$'s model of the world. The combined set of samples is $\mathbf{D} = \cup_i \mathbf{D}_i$ of size $M$. The corresponding empirical distribution is $\hat{p}$. The DM constructs an aggregate distribution $p$. The optimal aggregate distribution $p^*$ is posited to be the distribution the DM would learn from $\mathbf{D}$.

Since it is unrealistic to expect the DM to have access to the sources' sample sets, we consider how to use information about the sources' learned distributions to at least approximate $p^*$. Specifically, we consider the situation where the DM knows the sources' distributions and has a good estimate of the percentage $\alpha_i = M_i/M$ of the combined set of samples each source $i$ has observed as well as what learning method it used.

We make a number of assumptions. First, we assume that the samples are not noisy or otherwise corrupted, and they are complete (no missing values).

Second, we assume that the individual sample sets are disjoint (so $M = \sum_i M_i$). This implies that the concatenation of the $\mathbf{D}_i$ equals $\mathbf{D}$, so we don't have to concern ourselves with repeats when aggregating. This assumption is not always appropriate. It is invalidated when multiple sources observe the same event. However, there are interesting domains where this property holds. For example, in our motivating medical domain, doctors are likely to have seen disjoint sets of patients.

Third, we assume that the sources believe their samples to be IID — independent and identically distributed. The machine learning algorithms used in practice commonly rely on this assumption.

Finally, we assume that the samples in the combined set $\mathbf{D}$ are sampled from $\pi$ and IID. This assumption may appear overly restrictive at first glance. For one, it may seem to preclude the common situation where sources receive samples from different subpopulations. For example, if doctors are in different parts of the world, the characteristics of the patients they see will likely be different.

In fact, we can accomodate this situation within our framework by assuming $\pi$ is a distribution over the domain variables and a source variable $S$ which takes the different sources as values; $S = i$ means source $i$ observed the instantiated domain variables. This generalized distribution is sampled IID. Each $\mathbf{D}_i$ consists of the subset of samples where $S = i$. It is not necessary to keep around the $S$ values; computing the $p_i$ and $p^*$ without $S$ will give the same results as learning distributions over the complete samples and marginalizing out $S$. Thus, although samples will be IID, different subpopulation distributions will be possible, captured by different conditional probability distributions of the domain variables given distinct values of $S$.[1]

## 3 Aggregating Learned Joint Distributions

We first consider the case where sources have learned joint distributions, and the aggregate is also a joint.

### 3.1 Learning joint distributions: review

Given samples of a variable $X$, the goal of a learner is to estimate the probability of future occurences of each value of $X$. In our setting, the domain of $X$ is $\mathcal{W}$ and the parameters that need to be learned are the $|\mathcal{W}|$ probabilites. The distribution over $X$ is parameterized by $\Theta$. Two standard approaches are *Maximum Likelihood Estimation (MLE)* and *Maximum A Posteriori estimation (MAP)*.

[1] Two implications of this formulation are that the assumption that the $\mathbf{D}_i$ are disjoint is implicit and $\alpha_i$ will approach $\pi(S = i)$ as $M$ approaches $\infty$ for all $i$.

An MLE learner chooses the member of a specified family of distributions that maximizes the likelihood of the data:

**Definition 1** *If $X$ is a random variable,* $\mathrm{dom}(X) = \{x_1, \ldots, x_k\}$*, and* $\Theta = \langle \Theta_1, \ldots, \Theta_k \rangle$ *where* $\Theta_i = P(x_i \mid \Theta)$*, then the* MLE distribution *over $X$ given data set* $\mathbf{D}$ *is*

$$\mathrm{MLE}(X, \mathbf{D}) = \arg\max_{\theta} P(\mathbf{D} \mid \theta)$$

It is easy to show that the MLE distribution is the empirical distribution if samples are IID.

MAP learning, on the other hand, follows the Bayesian approach to learning which directs us to put a prior distribution over the value of any parameter we wish to estimate. We treat these parameters as random variables and define a probability distribution over them. More formally, we now have a joint probability space that includes both the data and the parameters.

**Definition 2** *If $X$ is a random variable,* $\mathrm{dom}(X) = \{x_1, \ldots, x_k\}$*, and* $\Theta = \langle \Theta_1, \ldots, \Theta_k \rangle$ *where* $\Theta_i = P(x_i \mid \Theta)$*, then the* MAP distribution *over $X$ given data set* $\mathbf{D}$ *and prior* $P(\Theta)$ *is the distribution*

$$\mathrm{MAP}(X, P(\Theta), \mathbf{D}) \equiv P(X \mid \mathbf{D}) = \int P(X \mid \Theta) P(\Theta \mid \mathbf{D}) \, d\Theta$$

The appropriate conjugate prior for variables with multinomial distributions is Dirichlet. $\mathrm{Dir}(\Theta \mid \gamma_1, \ldots, \gamma_k)$, where each $\gamma_i$ is a hyperparameter such that $\gamma_i > 0$.

We will assume that Dirichlet distributions are assessed using the *method of equivalent samples*: given a prior distribution $\rho$ over $X$ and an estimated sample size $\xi$, $\gamma_i$ is simply $\rho(x_i)\xi$. We use these to parameterize MAP:

**Definition 3** *If $X$ is a random variable,* $\mathrm{dom}(X) = \{x_1, \ldots, x_k\}$, $\Theta = \langle \Theta_1, \ldots, \Theta_k \rangle$ *where* $\Theta_i = p(x_i \mid \Theta)$*, $\rho$ is a probability distribution over $X$, and $\xi > 0$, then* $\mathrm{MAP}(X, \langle \rho, \xi \rangle, \mathbf{D})$ *denotes the distribution* $\mathrm{MAP}(X, p_0, \mathbf{D})$ *where* $p_0 = \mathrm{Dir}(\Theta \mid \rho(x_1)\xi, \ldots, \rho(x_k)\xi)$.

We will omit the $X$ argument from the MLE and MAP notation since it is understood.

### 3.2 LinOP: review

Let us turn to the problem of aggregation. We will show that joint aggregation essentially reduces to LinOP. LinOP was proposed by Stone in [Sto61], but is generally attributed to Laplace. It aggregates a set of joint distributions by taking a weighted sum of them:

**Definition 4** *Given probability distributions $p_1, \ldots, p_L$ and non-negative parameters $\beta_1, \ldots, \beta_L$ such that $\sum_i \beta_i = 1$, the LinOP operator is defined such that, for any $w \in \mathcal{W}$,*

$$\mathrm{LinOP}(\beta_1, p_1, \ldots, \beta_L, p_L)(w) = \sum_i \beta_i p_i(w).$$

LinOP is popular in practice because of its simplicity. As described in [GZ86], it also has a number of attractive properties such as *unanimity* (if all the $p_i = p'$, then LinOP returns $p'$), *non-dictatorship* (no one input is always followed), and the *marginalization property* (aggregation and marginalization are commutative operators). However, LinOP has often been dismissed in the aggregation communities as a normative aggregation mechanism, primarily because it fails to satisfy a number of other properties deemed to be necessary of any reasonable aggregator, e.g., the *external Bayesianity* property (aggregation and conditioning should commute) and the preservation of shared independences. Furthermore, typical approaches to choosing the weights are often criticized as being ad-hoc.

However, this dismissal may have been overly hasty. LinOP proves to be the operator we are looking for in our framework: using it is equivalent to having the DM learn from the combined data set under intuitive assumptions.

### 3.3 MLE aggregation

Suppose the sources and the DM are MLE learners. As has been known in statistics for some time, the DM need only compute the LinOP of the sources' distributions.

**Proposition 1 ([Win68, Mor83])** *If* $p_i = \mathrm{MLE}(\mathbf{D}_i)$ *for each* $i \in \{1, \ldots, L\}$ *and* $p^* = \mathrm{MLE}(\mathbf{D})$*, then* $p^* = \mathrm{LinOP}(\alpha_1, p_1, \ldots, \alpha_L, p_L)$.

Although straight-forward, this proposition is illuminating. For one, the weight corresponding to each source has a very clear meaning; it is the percentage of total data seen by that source. The DM only needs to provide accurate estimates of these percentages. A high weight indicates that the DM believes a source has seen a relatively large amount of data and is, hence, likely to be very reliable. Thus, we address a common criticism of LinOP, that the weights are often chosen in an ad-hoc fashion. Also, if $M$ is known, the DM can compute the number of samples in $\mathbf{D}$ that were $w$: $M\mathrm{LinOP}(\alpha_1, p_1, \ldots, \alpha_L, p_L)$. Thus, LinOP can be viewed as essentially storing the *sufficient statistics* for the DM learning problem.

It is now easy to see why a property such as preservation of independence will not always hold given our learning-based semantics. In our framework, sources do not have strong beliefs about independences; any believed independence depends on how well it fits the source's data. The independence preservation property does not take into account the possibility that, because of limited data, sources may all have learned independences which are not justified if all the data was taken into account.

Consider, for example, the following distribution over two variables $A$ and $B$: $\pi(ab) = 1/4$, $\pi(a\bar{b}) = 1/6$, $\pi(\bar{a}b) = 1/3$, and $\pi(\bar{a}\bar{b}) = 1/4$. Obviously, $A$ and $B$ are not independent. Suppose two sources have each received a set of six samples from this distribution: $\mathbf{D}_1$ consists of one each of $ab$ and $a\bar{b}$, two each of $\bar{a}b$ and $\bar{a}\bar{b}$; $\mathbf{D}_2$ consists of one each of $a\bar{b}$ and $\bar{a}\bar{b}$, two each of $ab$ and $\bar{a}b$. Further suppose each used MLE to learn a distribution over $A$ and $B$. $A$ and $B$ are independent in each of these distributions. The LinOP distribution, on the other hand, effectively takes into account the evidence seen by both sources and actually computes $\pi$ where the variables are not independent.

### 3.4 MAP aggregation

MLE learners are known to have problems with overfitting and low-probability events for which data never materialized. MAP learning often does a better job of dealing with these problems, especially when data is sparse.

Consequently, suppose the sources and the DM are MAP learners with Dirichlet priors. The optimal aggregate distribution is a variation on LinOP:[2]

**Proposition 2** *Suppose, for each* $i \in \{1,\ldots,L\}$, $p_i = \mathrm{MAP}(\langle\rho_i,\xi_i\rangle,\mathbf{D}_i)$ *and* $p^* = \mathrm{MAP}(\langle\rho,\xi\rangle,\mathbf{D}_i)$. *Then,*

$$p^*(w) = \frac{1}{M+\xi}\left(M\mathrm{LinOP}(\alpha_1,\mathrm{p}_1,\ldots,\alpha_L,\mathrm{p}_L) + \rho(w)\xi\right) + \sum_i \frac{\xi_i}{M+\xi}\left(p_i(w) - \rho_i(w)\right). \quad (1)$$

The first term in Equation 1 is the DM's MAP estimation, the second term accounts for the sources' priors by subtracting out their effect.

**Corollary 2.1** *Suppose, for each* $i \in \{1,\ldots,L\}$, $p_i = \mathrm{MAP}(\langle\rho_i,\xi_i\rangle,\mathbf{D}_i)$ *and* $p^* = \mathrm{MAP}(\langle\rho,\xi\rangle,\mathbf{D}_i)$. *Then,*

$$\lim_{\substack{\xi/M\to 0\\ \xi_i/M\to 0\,\forall i}} p^* = \mathrm{LinOP}(\alpha_1,\mathrm{p}_1,\ldots,\alpha_L,\mathrm{p}_L).$$

Thus, as $M$ becomes large, the LinOP distribution approaches $p^*$. This is not surprising since it is well-known that MLE learning and MAP learning with Dirichlet priors are asymptotically equivalent. The implication is that if $M$ is large, not only do we not need to know $M$ to aggregate, we do not need to know what priors the sources used either. And if we approximate the aggregate distribution by the LinOP distribution, this approximation will improve the more samples seen by the sources.

## 4 Aggregating Learned Bayesian Networks

Bayesian networks (BNs) are structured representations of probability distributions. A BN $b$ consists of a directed acyclic graph (DAG) $g$ whose nodes are the $N$ random variables. The parents of a node $X$ are denoted by $\mathrm{Pa}(X)$; $\mathrm{pa}(X)$ denotes a particular assignment to $\mathrm{Pa}(X)$. The structure of the network encodes marginal and conditional independencies present in the distribution. Associated with each node is the conditional probability distribution (CPD) for $X$ given $\mathrm{Pa}(X)$.

[2] We omit proofs for lack of space.

We consider the case where sources' beliefs are represented as BNs learned from data. We briefly review the techniques used for learning BNs from data. For a more detailed presentation, see [Hec96].

### 4.1 Learning Bayesian networks: review

If the structure of the network is known, the task reduces to statistical parameter estimation by MLE or MAP. In the case of complete data, the likelihood function for the entire BN conveniently decomposes according to the structure of the network, so we can maximize the likelihood of each parameter independently.

If the structure of the network is not known, we have to apply Bayesian model selection. More precisely, we define a discrete variable $G$ whose states $g$ correspond to possible models, i.e., possible network structures; we encode our uncertainty about $G$ with the probability distribution $P(g)$. For each model $g$, we define a continuous vector-valued variable $\Theta_g$, whose instantiations $\theta_g$ correspond to the possible parameters of the model. We encode our uncertainty about $\Theta_g$ with a probability distribution $P(\theta_g \mid g)$.

We score the candidate models by evaluating the *marginal likelihood* of the data set $\mathbf{D}$ given the model $g$, that is, the *Bayesian score* $P(\mathbf{D} \mid g) = \int P(\mathbf{D} \mid \theta_g, g)P(\theta_g \mid g)P(g)d\theta_g$.

In practice, we often use some approximation to the Bayesian score. The most commonly used is the MDL score, which converges to the Bayesian score as the data set becomes large. The MDL score is defined as

$$\mathrm{score}_{\mathrm{MDL}}(\mathrm{b}' : \mathbf{D}) = M\sum_{i=1}^{N}\sum_{\mathrm{pa}(X_i)}\sum_{x_i}\hat{p}(x_i,\mathrm{pa}(x_i))\log\hat{p}(x_i|\mathrm{pa}(x_i)) - \frac{\log M}{2}\mathrm{Dim}[\mathrm{g}'] - \mathrm{DL}(\mathrm{g}')$$

where $\mathrm{Dim}[\mathrm{g}']$ is the number of independent parameters in the graph and $\mathrm{DL}(\mathrm{g}')$ is the description length of $g'$. Finding the network structure with the highest score has been shown to be NP-hard in general. Thus, we have to resort to heuristic search. Since the search can easily get stuck in a local maximum, we often add random restarts to the process. The BN learning algorithm is presented in Figure 1.

Why are we interested in learning BNs rather than joint

```
1.  pick a random DAG g
2.  parameterize g to form b
3.  score b
4.  loop
5.      for each DAG g' differing from g by
        adding, removing, or reversing an edge
6.          parameterize g' to form b'
7.          score b'
8.      pick the b' with the highest score and replace
        g with g' and b with b' if score(b') > score(b)
9.  until no further change g
10. return b.
```

Figure 1: Bayesian network learning algorithm.

distributions? Besides some obvious reasons concerning compact representation and efficient inference, a distribution learned by the BN algorithm may be closer to the original distribution used to generate the data in the first place.

First, note that the networks which can be parameterized to represent exactly the MLE- or MAP-learned joint distributions are, in general, fully connected. Intuitively, a distribution learned from finite sample data will always be a little noisy, so true independences will almost always look like slight dependences mathematically. As a result, the BNs we are interested in (either for the sources or for the DM) will not be exact representations of the independencies present in the MLE- or MAP-learned distributions, but, rather, will account for this overfitting.

BN learning 'stretches' the distribution that best fits the data to match candidate network structures. For every structure, we look for the best (producing the highest score) parameterization *of that structure*. The score balances the fit to the data with model complexity.

### 4.2 LinOP-based Aggregation Algorithm

Now suppose each source has learned a BN $b_i$ with DAG $g_i$ from $\mathbf{D}_i$ using the MDL score and the DM is given these BNs as well as the $\alpha_i$. According to our semantics, the aggregate BN should be as close as possible to the one the DM would learn from $\mathbf{D}$.

We cannot apply the BN learning algorithm directly, since we don't have the data used by sources to learn their models. A simple solution would be to generate samples from each source model and train the DM on the combined set. That algorithm, although appealingly simple, raises some new questions. It is not clear how many samples we should generate from each source. One possibility would be to use the same number as the (estimated) number of samples that each source used to learn its model. However, if that number is small, the samples will not represent the generating distribution adequately, introducing additional noise to the process. If we generate more samples than each source saw (increasing it proportionally to preserve the $\alpha_i$ settings), we give too much weight to the MLE component of the score, thus possibly choosing a suboptimal network. In fact, our experiments described in Section 5 show that this algorithm does very badly in practice.

Instead, we can adapt the BN learning algorithm to use sources' distributions instead of samples.

The main difference is in the way we compute the MLE/MAP parameters for each structure we consider and the way we compute the score (lines 2, 3, 6 and 7 in Figure 1). Our algorithm relies on the observation that it is not necessary to have the actual data to learn a BN; it is sufficient to have their empirical distribution. As we have demonstrated in Section 3, we can come up with said distribution by applying the LinOP operator to distributions learned by our sources.

We can take advantage of the marginalization property of LinOP to make computation more efficient. As is noted in [PW99], we can parameterize the network in top-down fashion by first computing the distribution over the roots, then joints over the second layer variables together with their parents, etc. The conditional probabilities can be computed by dividing the appropriate marginals (using Bayes Law). In many cases, that would require only local computations in sources' BNs.

The MDL score also requires knowing only the empirical distribution for $\mathbf{D}$ and $M$. Again, since the empirical distribution is the LinOP distribution if the weights are chosen correctly and the sources used MLE or MAP (assuming sufficient data) learning, it is possible to score the candidate networks without having the actual data. Furthermore, the marginals used in the MLE score are family marginals. If the previous parameterization step is done by computing marginals, then these will have already been computed.

Although the MDL score requires knowledge of $M$, this dependence may not be strong, especially for large $M$ in which case the second term is dominated by the likelihood term and $M$ becomes a factor common to all networks and can be ignored. Otherwise, a rough approximation of $M$ should suffice.

As in traditional BN learning, caching can make the parameterization and scoring of 'neighboring' networks more efficient. Since we are making only local changes to the structure, only a few parameters will need updating. If an arc is added or removed, we only need to recompute new parameters for the child node, and if an arc is switched, we only need to recompute parameters for the two nodes involved. Also, since these LinOP marginals don't change, caching computed values may help to further speed up future computations.

# 5 Experiments

We implemented the BN aggregation algorithm in Matlab using Kevin Murphy's Bayes Net Toolbox[3] and explored its behavior by running experiments on the well-known, real-life Alarm network [BSCC89], a 37-node network used as part of a system for monitoring intensive care patients, and on the smaller 8-node artificial Asia network [LS88].

In our experiments, we learned two source BNs from data sampled from the original BN, then aggregated the results using our algorithm (AGGR). We had both the sources and the DM use MAP to parameterize their networks. In computing LinOP, we used the $\alpha_i$ as weights. We compared our proposal's accuracy against learning from the combined data sets (OPT) by plotting the Kullback-Leibler (KL) divergence [Kul59][4] of each distribution from the true distribution for different values of $M = |\mathbf{D}|$.

## 5.1 Sensitivity to $M$

We considered the situation where the DM knows the priors used by the sources and adjusts for the unduly large number of imaginary samples. All sources and DMs used the Dirichlet prior defined by the uniform distribution and an estimated sample size of 1. We varied the total number of samples $M$ between 200 and 20000, having sources see the same number of samples in some cases and different numbers in others. We conducted multiple runs for each setting and averaged them. Figure 2(a) plots the averages for the Alarm network when sources have equal $\alpha_i$. Due to software limitations, we had to start each structure search with the fully disconnected graph and used no random restarts for this larger network. As can be seen, in spite of the limited search, our algorithm does fairly well as far as coming close to the optimal and improving on the sources. Not surprisingly, the KL divergence drops as the total number of samples increases. Furthermore, the experiments on sources with different $\alpha_i$ showed no dependence of the performance of the algorithm on the relative difference in $\alpha_i$.

We ran similar experiments on Asia. Here, we varied the number of samples between 200 and 3000, with five runs per setting. For each run, we used five random restarts. Figure 2(b) plots the average for each setting. The plot shows that when we are able to explore the search space sufficiently in the learning and aggregation algorithms, our algorithm consistently improves on the sources and closely approximates to the optimal.

[3]Available at http://www.cs.berkeley.edu/ murphyk/bnt.html.

[4]The KL divergence of distribution $q$ from $p$ is defined as $\sum_{w \in \mathcal{W}} p(w) \log \frac{p(w)}{q(w)}$.

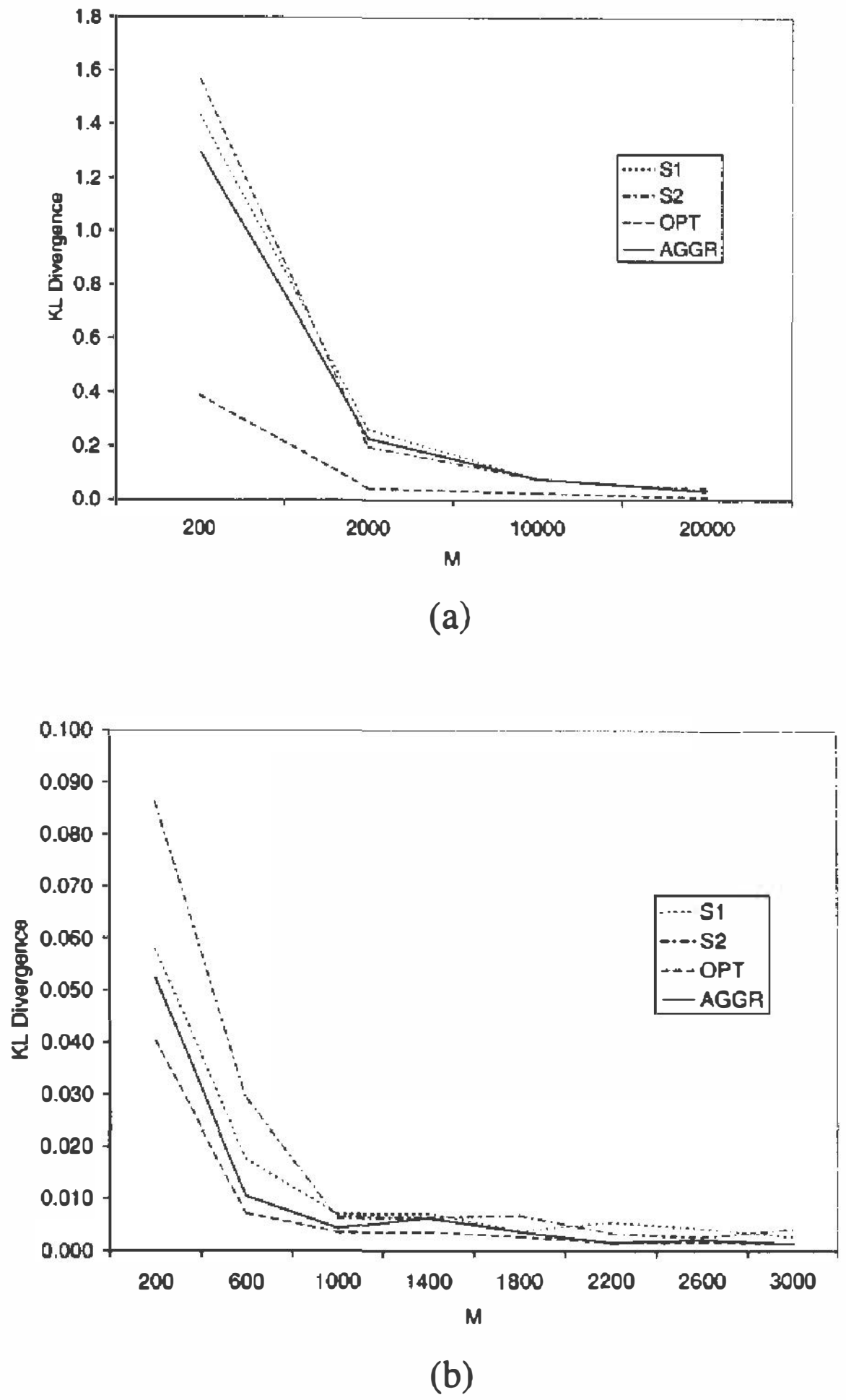

Figure 2: Sensitivity to $M$ (a) Alarm network results. (b) Asia network results.

## 5.2 Sensitivity to the DM's estimation of $M$

We hypothesized earlier that the actual value of the DM's estimate of M does not matter all that much. To demonstrate this, we ran experiments on the Asia network similar to those above, but leaving $M$ fixed and varying the DM's estimate 1 order of magnitude above and below $M$. Figures 3(a) summarizes the results for $M = 100$.

Any approximation above 0.25 orders of magnitude below $M$ provides improvement over the sources. Estimates below this made the complexity penalty sufficiently strong to select DAGs with fewer arcs than the original and underfit the data. On the other hand, although overestimating $M$ did not increase the KL distance from the original, there is a danger of extreme overestimates causing overfitting. However, we did not find any increase in the complexity of the aggregate networks for the 1 order of magnitude range we considered; they remained at 8–9 arcs on average.

Figure 3(b) summarizing the results for $M = 10000$ shows that, as predicted, the range of "slack" increases with $M$; the more samples seen by the sources, the less important the accuracy of the DM's estimate.

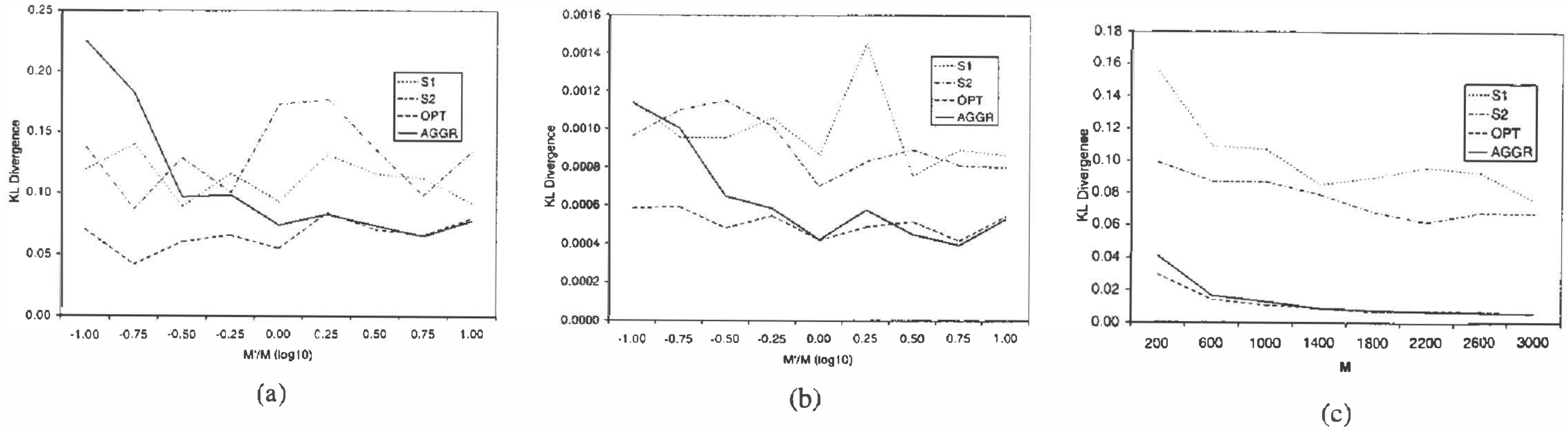

Figure 3: Asia network results (a) varying DM's estimate of $M$ ($M = 100$). (b) varying DM's estimate of $M$ ($M = 10k$). (c) with different subpopulations.

### 5.3 Subpopulations

Our algorithm performs well when combining source distributions learned based on samples from different subpopulations. To show this, we modified the Asia network to accomodate two sources, a doctor practicing in San Francisco and one practicing in Cincinnati. The probability distributions of the two root nodes in the Asia network, representing whether a patient smokes and whether she has visited Asia would be significantly different for the two doctors. A patient from San Francisco is less likely to be a smoker, and one from Cincinnati is less likely to have visited Asia. Thus, we added a source variable as described in Section 2, gave the sources equal priors of seeing patients, made the source variable a parent of the two root variables, and gave them appropriate CPDs. We drew $M$ samples from this extended network and had each source learn from the appropriate subset, then used AGGR to combine the results using the correct $\alpha_i$ and $M$. Figure 3(c) plots the KL divergence of each distribution from the original distribution with the source variable marginalized out. Because the sources are learning the distributions for different subpopulations, what they learn is relatively far from the overall distribution. The DM takes advantage of the information from both sources and learns a BN that approximates the original much more closely than either source.

### 5.4 Comparison to sampling algorithm

In each of the above experiments, we also compared the performance of our algorithm to the alternative intuitive algorithm SAMP we described in Section 4.2 in which we sample $\alpha_i M$ samples from each source $i$'s BN and learn a BN from the combined data. SAMP did very badly in general, consistently worse than not only AGGR, but worse than the sources as well, often by an order of magnitude.

## 6 Related Work

A wealth of work exists in statistics on aggregating probability distributions. Good surveys of the field include [GZ86, CW99]. Many of the earlier, axiomatic approaches suffered from a lack of semantical grounding. For this reason, the community moved towards modeling approaches instead. The most studied approach has been the *supra-Bayesian* one, introduced in [Win68] and formally established in [Mor74, Mor77]. Here, the DM has a prior not only over the variables in the domain, but over the possible beliefs of the sources as well. She aggregates by using Bayesian conditioning to incorporate the information she receives from the sources. In fact, Proposition 1 derives from this body of work. However, almost all of this work has been restricted to aggregating beliefs represented as point probabilities or odds, or joint distributions.

There has been some recent interest, particularly in AI, in the problem of aggregating structured distributions including [MA92, MA93, PW99]. But, like the early axiomatic approaches in statistics, much of this work focuses on attempting to satisfy abstract properties such as preserving shared independences, and often runs into impossibility results as a consequence.

In some sense, what we are doing could also be viewed as ensemble learning for BNs. Ensemble learning involves combining the results of different weak learners to improve classification accuracy. Because of its simplicity, LinOP is often used without justification to do the actual combination. Our results justify this use when the weak learners use MLE, MAP, or BN learning.

Another new area in AI that bears similarities to our work is that of on-line or incremental learning of BNs (e.g.,

[Bun91, LB94, FG97]). There, we are given a continuous stream of samples and we want to maintain a BN learned from all the data we have seen so far. Because the stream is very long, it is generally not possible to maintain the full set of sufficient statistics. Approaches range from approximating the sufficient statistics to restricting the network that can be learned. We essentially do the former by assuming that the sufficient statistics for the data seen by each source is encoded in its network. Cross-fertilization between the two fields may prove profitable.

## 7 Conclusion

We have presented a new approach to belief aggregation. We believe that we cannot formulate that problem precisely or measure success of different techniques without answering questions about the way in which sources' beliefs were formulated. We argued that a framework in which the sources are assumed to have learned their distributions from data is both intuitively plausible and leads to a very natural formulation of the optimal DM distribution — one which would be learned from the combined data sets — and a natural success measure — a distance from the generating, 'true' distribution.

Based on the observation that LinOP is the appropriate operator for this framework if sources and DM are MLE learners, we presented a LinOP-based algorithm to aggregate beliefs represented by Bayesian networks. Our preliminary results show that this algorithm performs very well.

One direction of future work will involve finding ways to relax the various assumptions. For example, we would like to extend the framework to allow for continuous variables and to allow for dependence between sources' sample sets.

In our framework, the DM completely ignores sources' priors. This may be appropriate if the priors are known to be unreliable or uninformative. However, the priors used in real applications are often informative in and of themselves. Thus, a second direction will involve finding valid ways of taking advantage of sources' priors to improve the quality of the aggregation. For example, if sources use Dirichlet priors and the DM trusts their estimated sample sizes, she may chose to incorporate them into her estimate of $M$.

### Acknowledgements

Pedrito Maynard-Reid II was partially supported by a National Physical Science Consortium Fellowship. Urszula Chajewska was supported by the Air Force contract F30602-00-2-0598 under DARPA's TASK program.